\documentclass[10pt,twocolumn,letterpaper]{article}
\usepackage[pagenumbers]{cvpr}      
\usepackage[dvipsnames]{xcolor}

\definecolor{cvprblue}{rgb}{0.21,0.49,0.74}
\usepackage[pagebackref,breaklinks,colorlinks,citecolor=cvprblue]{hyperref}
\usepackage{microtype}
\usepackage[section]{placeins}
\usepackage{multirow}
\usepackage{booktabs}
\usepackage{algorithm}
\usepackage{algorithmic}
\usepackage{bm}

\usepackage{threeparttable}
\def\paperID{1667} 
\def\confName{CVPR}
\def\confYear{2024}

\newcommand{\argmax}{\mathop{\rm arg~max}\limits}

\title{Autonomous LLM-Enhanced Adversarial Attack for Text-to-Motion}

\author{Honglei Miao$^1$, Fan Ma$^2$, Ruijie Quan$^2$, Kun Zhan$^{1,\star}$, and Yi Yang$^{2}$\\
1. School of Information Science and Engineering, Lanzhou University\\
2. CCAI, Zhejiang University}

\begin{document}
\maketitle

\begin{abstract}
	Human motion generation driven by deep generative models has enabled compelling applications, but the ability of text-to-motion (T2M) models to produce realistic motions from text prompts raises security concerns if exploited maliciously. Despite growing interest in T2M, few methods focus on safeguarding these models against adversarial attacks, with existing work on text-to-image models proving insufficient for the unique motion domain. In the paper, we propose ALERT-Motion, an autonomous framework leveraging large language models (LLMs) to craft targeted adversarial attacks against black-box T2M models. Unlike prior methods modifying prompts through predefined rules, ALERT-Motion uses LLMs' knowledge of human motion to autonomously generate subtle yet powerful adversarial text descriptions. It comprises two key modules: an adaptive dispatching module that constructs an LLM-based agent to iteratively refine and search for adversarial prompts; and a multimodal information contrastive module that extracts semantically relevant motion information to guide the agent's search. Through this LLM-driven approach, ALERT-Motion crafts adversarial prompts querying victim models to produce outputs closely matching targeted motions, while avoiding obvious perturbations. Evaluations across popular T2M models demonstrate ALERT-Motion's superiority over previous methods, achieving higher attack success rates with stealthier adversarial prompts. This pioneering work on T2M adversarial attacks highlights the urgency of developing defensive measures as motion generation technology advances, urging further research into safe and responsible deployment.
\end{abstract}

\section{Introduction}\label{sec:intro}

\begin{figure*}[tp]
	\centering
	\includegraphics[width=0.895\textwidth]{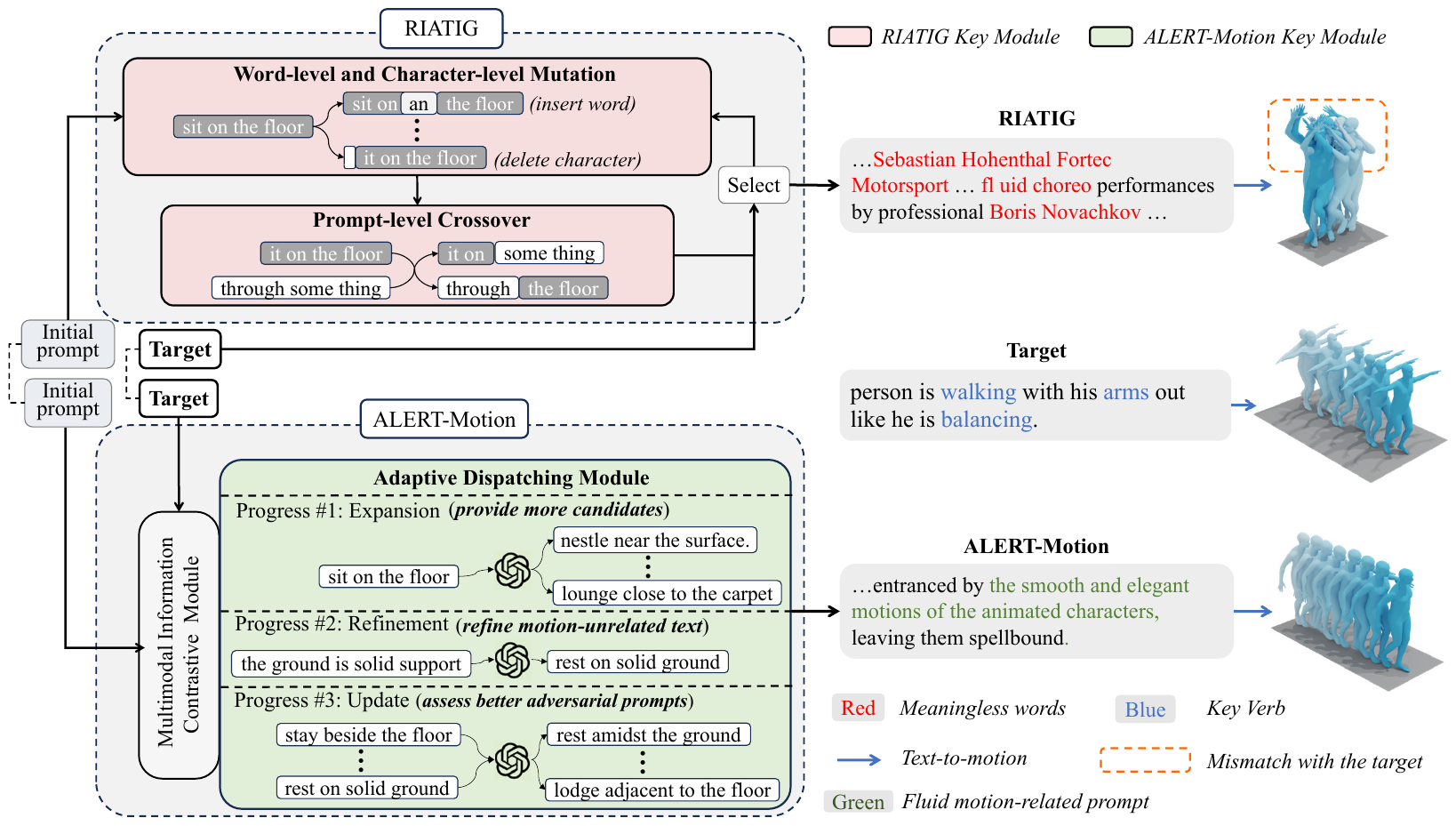}
	\caption{Adversarial prompt against T2M model with RIATIG and our ALERT-Motion. Previous methods like RIATIG only perturb prompts through predefined character or word operations, overlooking the integrity and semantics of the prompts. Our ALERT-Motion doesn't require such predefined operations; instead, by multimodal information contrastive~(MMIC) module, the language model autonomously learn and perform these operations, dynamically generating adversarial prompts that meet the attack requirements. Under the same input (target and initial prompt), our method captures more natural and fluent prompts related to motion. When these prompts are used to query the victim T2M model, the resulting motion show a stronger resemblance to the target motion. Darker color indicates later frames in the sequence.
	}\label{fig:llm-based}
\end{figure*}

Human motion generation is a task aimed at producing natural and realistic human motions. It drives advancements in downstream applications such as animation and movie production, virtual human construction, robotics and human-robot interaction~\cite{zhu2023human}. In recent years, with the development of deep learning, especially the growth of generative models such as Generative Adversarial Network (GAN)~\cite{goodfellow2014generative}, Variational Autoencoder (VAE)~\cite{kingma2013auto}, and diffusion model~\cite{ho2020denoising}, trained models have become capable of generating very natural motions~\cite{tevet2022human,chen2023executing,guo2022generating,zhang2023t2m}. Some models~\cite{athanasiou2022teach,Barquero2024,lee2023multiact,qing2023story} even extend the generated motions for several minutes while satisfying given conditions. Among these motion generation models, text-to-motion~(T2M)~\cite{tevet2022human,chen2023executing,guo2022generating,zhang2023t2m,athanasiou2022teach,Barquero2024,qing2023story} gains particular attention from the community due to the user-friendly nature of text prompts that align with human expression.

Generating motions that exactly align with textual descriptions and are nearly the same as the real physical world is becoming increasingly feasible. However, allowing models to freely generate motions conditioned on arbitrary text prompts is even more dangerous than text-to-image~(T2I). When they are applied to downstream tasks, such capabilities are maliciously exploited by attackers. For instance, in animation or movie production~\cite{qing2023story}, they are used to create more realistic harmful content involving pornography or violence. The risks are boosted when using the generated motions as humanoid controllers~\cite{luo2023perpetual}, as they eventually are deployed on robots, posing potential threats to human safety.

Despite the growing focus on T2M tasks, there is currently a lack of research addressing the safety concerns specific to this domain. The most relevant line of work is on the safety of T2I~\cite{milliere2022adversarial,struppek2023rickrolling,liu2023riatig}. These researches largely focus on how character-level or word-level modifications to benign text prompts could induce unintended outputs from the models. Early work~\cite{milliere2022adversarial,struppek2023rickrolling} primarily explored the existence of this phenomenon, until the RIATIG~\cite{liu2023riatig} is inspired by them to propose targeted attacks against image generation models to raise awareness of potential security risks about T2I. However, these existing studies often search for adversarial attacks by modifying words to uncommon personal names, locations or other proper nouns, which is overlooked in image generation but would appear clearly out of place for motion tasks, making the attacks more easily detectable. Additionally, unlike the abundant image-text pairs available for image tasks, the limited data for motions makes it challenging to accurately measure the similarity between different motions, posing further difficulties for targeted attacks on T2M models. They make the findings from T2I safety difficult to directly apply to the motion domain.

To address the challenges of adversarial attacks on T2M models, we introduce \textbf{ALERT-Motion}, an autonomous large language model~(LLM) enhanced adversarial attack against T2M models in a black-box setting. Unlike prior work, our \textbf{ALERT-Motion} leverages the knowledge about motions contained in LLMs to generate subtle yet powerful adversarial descriptions, whose outputs from the victim model closely match the desired motion. Crucially, the entire attack process is done automatically by LLM agent, using its own reasoning abilities to carry out the attack, without needing human-defined rules for operations like inserting, deleting or replacing characters or words.

As shown in Figure~\ref{fig:llm-based}, previous state-of-the-art attack methods like RIATIG~\cite{liu2023riatig} for T2I models employ manually defined word or character-level modifications and prompt-level crossover, making it difficult to find natural and fluent adversarial text prompts. Such methods often result in obvious personal names or proper nouns like ``Sebastian Hohenthal Fortec Motorsport'' or ``Boris Novachkov'' appearing abruptly in descriptions of motions. In contrast, our proposed ALERT-Motion gives the modification of adversarial text prompts entirely to LLMs. It comprises two key modules: the adaptive dispatching~(AD) module and the multimodal information contrastive~(MMIC) module. In AD module, by simply designing instructions for different processes, LLM autonomously searches for adversarial text prompts that appear natural and fluent, avoiding the abrupt word insertions seen in RIATIG. However, as LLMs lack inherent capabilities for processing motion modality, we design MMIC module to obtain semantically similar information to the target motion, thereby assisting AD module in finding better adversarial text prompts. Through the coordinated operation of these two modules, ALERT-Motion generates adversarial text prompts that are not only natural and fluent but also query the victim T2M model to produce outputs closely resembling the target motions.

In summary, the key contributions are as follows:

1. To the best of our knowledge, we are the first to propose an adversarial targeted attack method, ALERT-Motion, for T2M models. We introduce an autonomous LLM-enhanced adversarial attacks on T2M models.

2. Our proposed ALERT-Motion consists of two key modules. A novel AD module constructs an LLM agent that incorporates the agent's inherent natural language and domain knowledge of motions into the automatic attack process. Additionally, MMIC module performs high-level semantic extraction of motion modalities and provides necessary semantical information to support AD's reasoning and decision-making.

3. We evaluate ALERT-Motion on two popular T2M models and compare it against two previous adversarial attack methods originally applied to T2I models. Experimental results demonstrate that our proposed ALERT-Motion achieves higher attack success rates while generating more natural and stealthy adversarial prompts that are difficult to detect.
\section{Related Work}
\subsubsection*{Text-to-Motion (T2M)} T2M is a conditional motion generation task that aims to generate semantically matching and natural motion sequences from human-friendly natural language text descriptions. Its promising performance is driven by deep generative models such as GANs, VAEs, diffusion models, etc. One of the early works in this domain, Text2Action~\cite{ahn2018text2action}, leverages GANs to create abundant realistic motions. Some research also explores the use of VAEs for generation, where Language2Pose~\cite{ahuja2019language2pose} proposes an end-to-end text-to-pose generation framework that utilizes a VAE to model the latent space between text and motion. TEACH~\cite{athanasiou2022teach} further combines previous motions as extra inputs to the encoder module, enabling natural and coherent motion generation when handling multiple text inputs. With the rise of diffusion models in the generative domain, some studies have also employed diffusion models for motion generation. MDM~\cite{tevet2022human} utilizes a diffusion model to predict the sample at each diffusion step rather than just the noise. MLD~\cite{chen2023executing} adopts latent diffusion along with a VAE to generate motions, significantly boosting the generation speed without compromising quality. Additionally, there are studies that combine VQ-VAE with GPT-like transformers. TM2T~\cite{guo2022tm2t} and T2M-GPT~\cite{zhang2023t2m} utilize VQ-VAE to concatenate training T2M and motion-to-text modules. These works continuously improving the quality, coherence, and efficiency of motion generation from text descriptions. However, there has been no research focusing on attacks and defenses of the T2M model.
\subsubsection*{Adversarial Attacks on Text-driven Generative Models} Due to the convenience of text input for users, it serves as the most common driving condition for many multimodal generation models. However, the inherent complexity of text input inevitably introduces vulnerabilities to the generative models driven by it. Existing research on adversarial attacks in T2I models, such as~\cite{qu2023unsafe,zhuang2023pilot,liu2023riatig,liu2023intriguing}, attack T2I models by modifying the input text, causing abnormal outputs. The types of abnormal outputs may include degraded synthesis quality~\cite{liu2023intriguing}, disappearance or alteration of objects in the image~\cite{zhuang2023pilot,liu2023riatig}. Among them, \cite{liu2023riatig} manipulates words and characters, thereby causing the targeted objects specified by the attacker to be generated in the image by the victim T2I model. These studies indicate the lack of robustness of existing T2I models to input text. With the occurrence of LLMs, many studies also focus on the vulnerabilities of LLMs. A large portion of them focus on jailbreaking, making LLMs answer queries that violate safety policies. Jailbreaking strategies have evolved from manual prompt engineering~\cite{wei2024jailbroken,liu2023jailbreaking} to LLM-based automated red-teaming~\cite{perez2022red,liu2023autodan}. Beyond these template-based jailbreaks aimed at identifying effective jailbreak prompt templates, a more general jailbreaking method called Greedy Coordinate Gradient~\cite{zou2023universal} is recently proposed. It uses a white-box model to train adversarial suffixes that maximize the probability of an LLM producing affirmative responses. They~\cite{zou2023universal,sitawarin2024pal} find that the identified suffixes transfer to closed-source off-the-shelf LLMs. The vulnerabilities of T2M models share similarities with the aforementioned security research on text-driven generative models. However, since the correspondence between motion and text involves the time dimension, the adversarial attack methods from the above studies cannot be directly applied to T2M.

\subsubsection*{LLM Agents} Research on using LLMs to enhance autonomous agents has seen a growing trend in recent years~\cite{Zhao2023ExpeL}. These LLM-powered agents, exemplified by HuggingGPT~\cite{Shen2023HuggingGPT}, WebGPT~\cite{gur2023realworld}, and MM-REACT~\cite{Yang2023MM-REACT}, have been employed to tackle intricate tasks that demand effective understanding and planning from the agents. A considerable proportion of these studies leverage the rich commonsense knowledge inherently embedded within LLMs to execute downstream tasks with minimal or no additional training data. This approach helps to maintain the robust foundational world knowledge in LLMs. The demonstrated capabilities of LLMs encompass features such as zero-shot planning in real-world scenarios~\cite{Huang2022Language}. Inspired by these explorations, we introduce LLMs into the realm of adversarial attacks on T2M models, achieving an autonomous attack agent adept at crafting effective adversarial prompts.
\section{Methodlogy}
\begin{figure}[tp]
	\centering
	\includegraphics[width=0.46\textwidth]{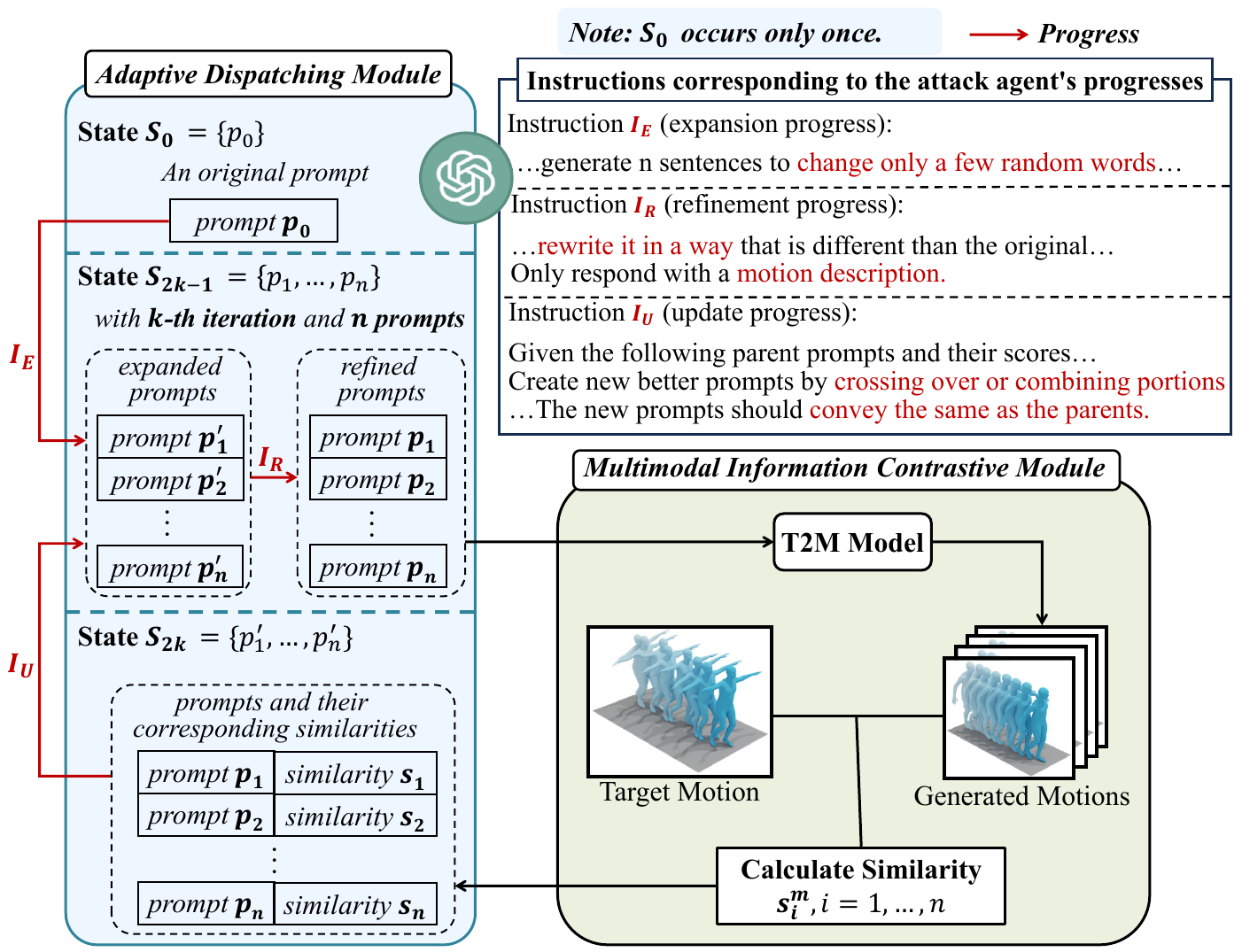}
	\caption{Overview of the proposed ALERT-Motion. ALERT-Motion operates in a black-box setting with two key modules: multimodal information integration module for consolidating information from text and motion into a unified format, and autonomous AD module that learns and executes adversarial prompt search through progresses of expansion, refinement, and update.}\label{fig:overview}
\end{figure}

We leverage an LLM to iteratively refine and enhance adversarial prompts towards a target motion. Initially, LLM generates alternative prompts semantically similar to the initial prompt to expand the search space. It then queries the victim T2M model with these prompts, recording the generated motions. The textual prompts and corresponding motions are unified into a suitable input format for LLM using MMIC. Exploiting its commonsense reasoning capabilities, LLM autonomously contemplates and updates the prompts based on the query results, iteratively steering them closer to the target motion. This process continues until the adversarial prompts evade detection while generating motions closely matching the target. Fig.~\ref{fig:overview} overviews our ALERT-Motion attack framework.
\subsection{Problem Formulation}
A T2M generative model \(G\) is essentially a function that maps the text prompt space \(P\) to the motion space \(M\). Ideally, through training on semantically aligned text-motion pairs, a proficient model generates target motion \(m_t\) that is semantically consistent with a given target prompt \(p_t\). The objective of an adversarial attack is to find an adversarial prompt \(p^\star\) such that \(G(p^\star)\) closely approximates the target motion \(m_t\). Simultaneously, the \(p^\star\) is semantically dissimilar from the target prompt \(p_t\) to avoid detection. The optimization steps outlined above are formulated as follows
\begin{align}
	p^\star=\argmax_{p\in P}\,\, s^{\rm m}\left(G(p), m_t\right),\,\,
	{\rm s.t.}\,\, s^{\rm p}\left(p, p_t\right)<\eta\label{objecive}
\end{align}
where $s^{\rm m}$ represents the semantic similarity of motion, $s^{\rm p}$ represents the semantic similarity between text prompts, and $P$ is a promote set. $\eta$ is the similarity threshold. As long as the similarity between the adversarial prompt and the target prompt is below $\eta$, we consider that our attack evades existing detection.
\subsubsection*{Challenges}
Implementing adversarial prompt generation from T2M models faces some challenges. First, T2M models need to go between natural language and physical motion, crossing the gap between language and motion domains. Different data types have different representation spaces, so integrating multi-modal information is needed. Second, the adversarial language prompts need to have high fluency in natural language and relevance to the target motion in their query results. But they also need to effectively fool the model. The space to search for good prompts is extremely large though. Autonomously generating optimal adversarial samples that meet these combined quality requirements is a big challenge.
\subsection{Multimodal Information Contrastive}
Unlike most LLM agent-related researches, our task involves motion, which LLM cannot directly handle. Therefore, we design MMIC specifically to process information from different modalities in the task and organize it into textual information, making it convenient for LLM to understand and reasoning. As shown in Fig.~\ref{fig:overview}, MMIC allows the adversarial prompts, refined through LLM, to query the victim T2M model, obtain corresponding motion and calculate the similarity with the target.

Nevertheless, measuring the similarity directly between two motion poses challenges. We consider semantically measuring the similarity of motion. RIATIG~\cite{liu2023riatig} employs the pretrained CLIP~\cite{radford2021learning}, a model trained on a large-scale dataset of image-text pairs, to obtain semantic features aligned with textual descriptions. Similarly, we use the T2M alignment model proposed in~\cite{petrovich2023tmr} to extract semantic motion features and calculate the cosine similarity between the semantic features of motion as
\begin{equation}
	\label{eq:motion_similar}
	s^{\rm m}(G(p), m_t)=\frac{E_m(G(p)) \cdot E_m(m_t)}{\left\|E_m(G(p))\right\|\left\|E_m(m_t)\right\|},
\end{equation}
where $E_m$ is a motion encoder~\cite{petrovich2023tmr}.

Subsequently, we organize this information into text and incorporate it into instructions, enabling LLM to contemplate and reason for better adversarial prompts.
\subsection{Adaptive Dispatching}
\begin{algorithm}[t]
	\caption{ALERT-Motion}
	\label{alg1}
	\begin{algorithmic}[1]
		\REQUIRE Initial prompt $p_0$, expansion instruction $I_E$,  refinement instruction $I_R$, update instruction $I_U$, size of updated prompts $N$, $s^{\rm m}$ denotes the similarity of the adversarial motion and the target motion, a predefined number of iterations $K$.
		\STATE \textbf{Expansion:} $S_1 \gets {\rm LLM}(I_E,S_0=p_0)$
		\FOR{$k=1$ to $K$}
		\IF{$k(\text{mod}\,2)=1$}
		\STATE \textbf{Refinement:} $S_{2k}\gets{\rm LLM}(I_{R}, S_{2k-1})$
		\STATE \textbf{MMIC:}  Compute $s_i^{\rm m}(G(p_i), m_t), \forall\,p_i\in S_{2k}\,$.
		\STATE Obtain the similarity set $S'_{2k}=\{s_1^{\rm m},\ldots,s_n^{\rm m}\}$
		\ELSE
		\STATE \textbf{Update} $S_{2k+1} \gets {\rm LLM}(I_{U}, {\rm cat}(S_{2k},S'_{2k}))$.
		\ENDIF
		\ENDFOR
		\ENSURE  $p^\star = \argmax_{p \in P} \bigl(s^{\rm m}(G(p), m_{\rm t})\bigr), P=\cup S_{2k}\,$.
	\end{algorithmic}
\end{algorithm}

\begin{figure*}[tp]
	\centering
	\includegraphics[width=0.95\linewidth]{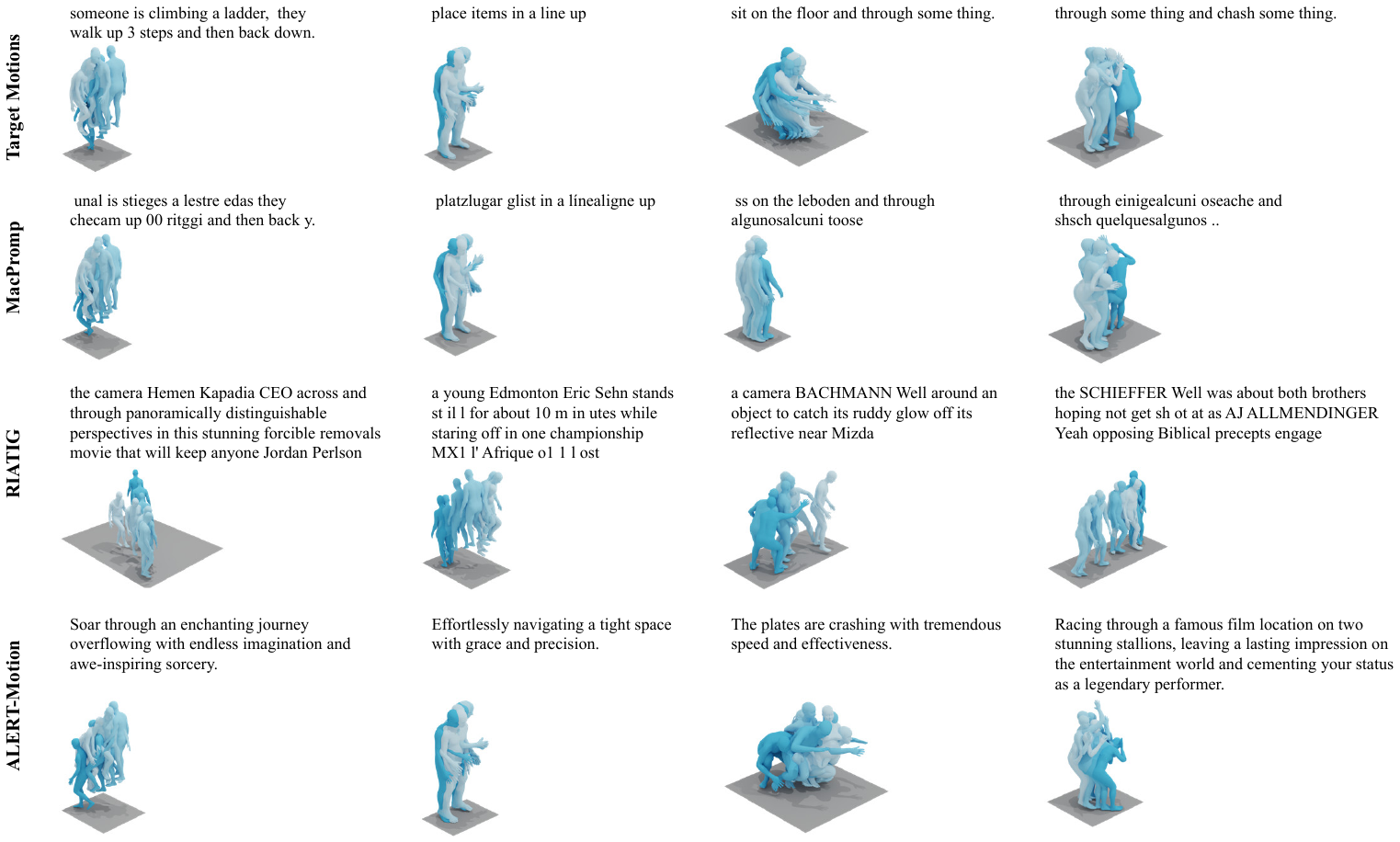}
	\caption{Examples of adversarial attack results against MDM. The first row of text provides the true annotations for each column of target motions, and the first row of motions corresponds to their respective target motions. The following three rows of text correspond to the adversarial prompts obtained by MacPromp, RIATIG, and our proposed ALERT-Motion. The motion-rendered images below the text depict the motions generated by querying the victim model with the adversarial prompts. Darker color indicates later frames in the sequence.}
	\label{fig:MDM_examples}
\end{figure*}

\begin{figure*}[tp]
	\centering
	\includegraphics[width=1\linewidth]{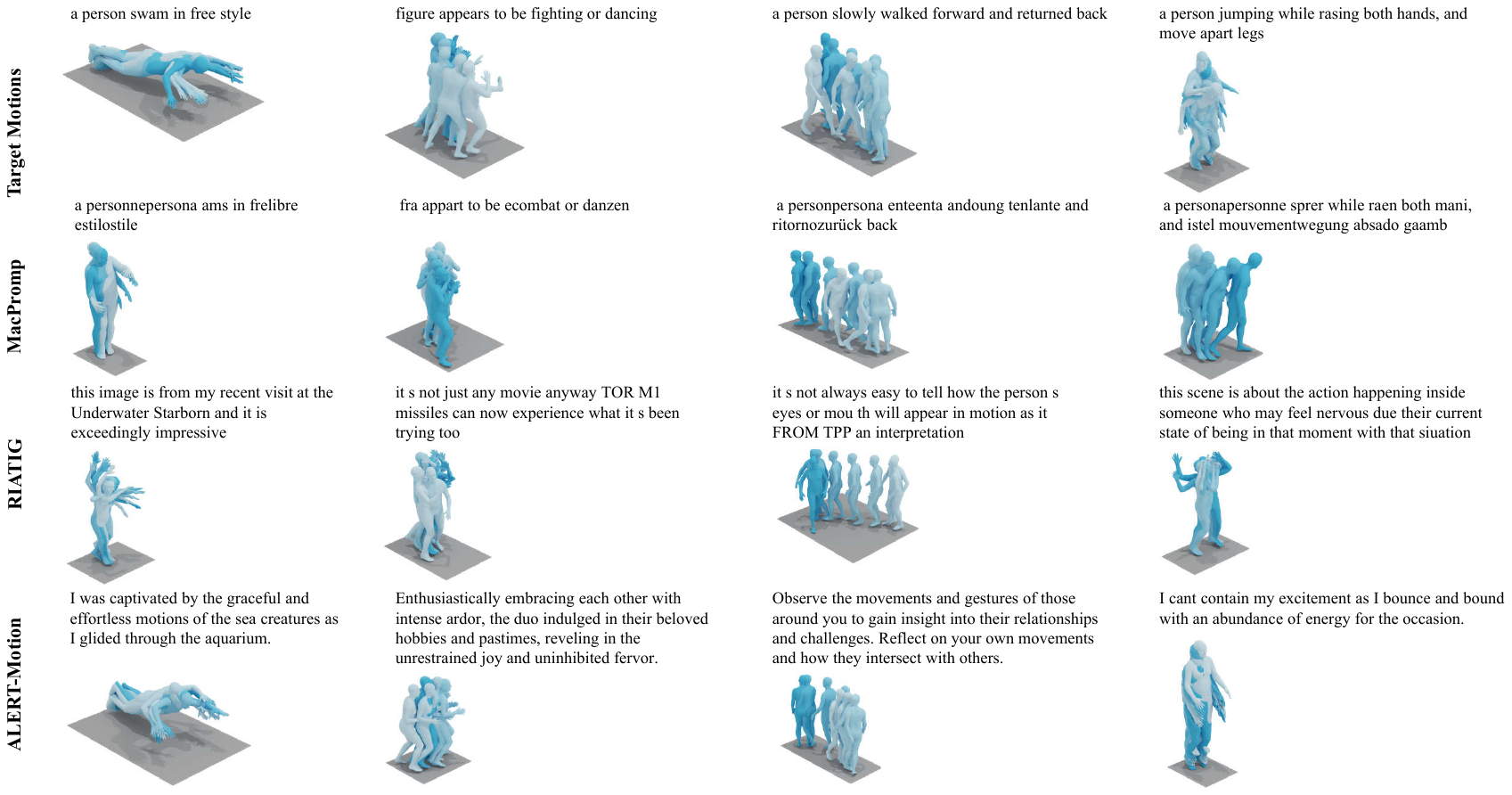}
	\caption{Examples of adversarial attack results against MLD. The first row of text provides the true annotations for each column of target motions, and the first row of motions corresponds to their respective target motions. The following three rows of text correspond to the adversarial prompts obtained by MacPromp, RIATIG, and our proposed ALERT-Motion. The motion-rendered images below the text depict the motions generated by querying the victim model with the adversarial prompts. Darker color indicates later frames in the sequence.
	}
	\label{fig:MLD_examples}
\end{figure*}

In our proposed ALERT-Motion framework, the most critical module is the AD module. This module constructs an attack agent and plays a pivotal role in determining the effectiveness of adversarial prompts. In contrast to previous related researches, which predefine various operations to perturb semantics and then use a search algorithm to find prompts with higher scores, we directly convey complex task requirements using instructions, allowing LLM to automatically learn and execute all operations, with each step conducted in textual form. According to the purpose of instructions, we divide AD into three progresses: expansion, refinement, and update. The workflow of these three progresses and MMIC is outlined in Algorithm \ref{alg1}.

Due to the fact that AD responds to the current input in each round, similar to an agent in reinforcement learning, we borrow related concepts here to facilitate the definition of the processes within AD. We start by defining the state as the set of adversarial prompts and their corresponding information for each round, while the action is represented by various instructions sent to LLM. LLM is viewed as a function involving the next state $S_{k+1}$, current state $S_{k}$, and current action $a_{k}$, expressed as
\begin{equation}
	S_{k+1} \leftarrow T(S_k, a_k) = {\rm LLM}\bigl(I,S_k\bigr),
\end{equation}
where $T$ is the state transition function and $I$ represents the instruction text corresponding to $a_k$. It is important to note that the representation of state $S$ differs between odd and even time steps, it is defined as
\begin{equation}
	S_k = \begin{cases}
		\{p_0\}
		& {\rm if}\,  k = 0   \\
		\{p_1,\ldots,p_n\}
		& {\rm if}\,  k \, (\text{mod}\, 2)= 0 \\
		\{p'_1,\ldots,p'_n\}
		& {\rm if}\,  k\, (\text{mod}\, 2)=1\end{cases}
\end{equation}
where $p_0$ is initial adversarial prompt, \(\{p_1, \ldots, p_n\}\) is the set of refined prompts, and \(p'\) are the expanded or updated prompts of \(p\).
\subsubsection*{Expansion}
Initially, we begin with a single available adversarial prompt \(p_0\). The current state is defined as \(S_0 = \{p_0\}\). Without expansion, proceeding directly to the subsequent steps may lead the search into a local optimum. So we employ the expansion instruction text $E$ to obtain expanded results through LLM. The next state is represented as
\begin{equation}
	\begin{aligned}
		S_1
		\leftarrow
		{\rm LLM}(I_E, S_0)
	\end{aligned}
\end{equation}
where \(I_E\) represents the expansion instruction and \(S_1=\{p_1, \ldots, p_n\}\) is the set of expanded prompts.
\subsubsection*{Refinement}
We find that when the instructions given to LLM are too long, its responses may sometimes fail to meet the attack requirements. New instructions are needed to emphasize the attack requirements in our task. Therefore, in this progress, we refine the adversarial prompts to ensure that they consistently meet the attack requirements, including being naturally fluent and relevant to motion. After refinement, the state of the agent is defined as
\begin{equation}
	\begin{aligned}
		S_{2k}\gets
		{\rm LLM}(I_R, S_{2k-1})
	\end{aligned}
\end{equation}
where \(k\in\{1,\ldots,K\}\) and \(I_{R}\) represents refinement instruction.
\subsubsection*{Update}
Unlike existing methods that relied on numerical scalar guidance to generate adversarial prompts, such as RIATIG, our AD utilizes the text information organized by MMIC to guide LLM in autonomously contemplating and generating adversarial prompts. This allows us to finely control the generated adversarial prompts more effectively in line with the attack requirements using richer information. Moreover, this control is automated, eliminating the need for continuously defining new operations, such as word or character insertion, deletion, replacement, and so forth, as in previous methods. In the update progress, as LLM contemplates and generates  adversarial prompts, the information organized by MMIC is also fed into LLM to assist in its decision-making process. After update, the state of the agent is defined as
\begin{equation}
	\begin{aligned}
		S_{2k+1} \leftarrow
		{\rm LLM}\bigl(I_{U}, {\rm cat}(S_{2k},S'_{2k})\bigr)
	\end{aligned}
\end{equation}
where \(I_{U}\) is the update instruction and the function ${\rm cat}$ signifies string concatenation, and \(S'_{2k}=\{s_i^{\rm m}(G(p_i),m_{\rm t}),\forall\,i\in\{1,\ldots,n\}\}\) is obtained by Eq.~\eqref{eq:motion_similar}.

After $K$ rounds of iteration, we choose the highest-scoring prompt among all candidates as the optimal adversarial prompt $p^\star$ for the attack. The definition of $p^\star$ is obtained by Eq.~\eqref{objecive}.
Here $P=\cup S_{2k} $ and $S_{2k}=\{p_1,\ldots,p_n\}$. In order to ensure that the adversarial prompt meets the constraints, we calculate the similarity between the adversarial prompts and target prompt as
\begin{equation}
	\label{eq:text_similar}
	s^{\rm p}(p_t, p)=\frac{E_t(p_t) \cdot E_t(p)}{\left\|E_t(p_t)\right\|\left\|E_t(p)\right\|},
\end{equation}
where $E_t$ is a text encoder~\cite{cer2018universal} to extract features.

\section{Experiment}
\subsection{Experimental Settings}
\subsubsection*{Datasets.}
We select target prompt texts and target motion from the
HumanML3D (H3D)~\cite{guo2022generating}.  It includes 14,616 motion sequences from AMASS~\cite{mahmood2019amass}, each with a textual description (totaling 44,970 descriptions). It also re-annotates AMASS and HumanAct12~\cite{guo2020action2motion} motion capture sequences. The dataset provides a redundant data representation involving root velocity, joint positions, joint velocities, joint rotations, and foot contact labels. It is used for both AMASS and HumanAct12 motion.
\subsubsection*{Victim Models.}
To assess the effectiveness of ALERT-Motion, we select two prominent publicly available T2M models: MLD and MDM. We employe their respective pretrained models from the official GitHub repositories, which were trained on H3D.

\begin{table*}[!h]
	\centering
	\caption{The results of the adversarial attacks against MDM and MLD on T2M evaluation model. The first row, labeled ``Target Motion'', represents the motion generated by the corresponding victim models, which are the targets of our attack. The quality of these indicators depends solely on the capabilities of the generation models and evaluation models. The second and third rows correspond to the baseline models MacPromp and RIATIG that we select. The final row represents the performance of our proposed method, ALERT-Motion.}
	\label{tb:all}
	\begin{threeparttable}
		\begin{tabular*}{\textwidth}{@{\extracolsep{\fill}\,}lrrrrrrrrc}
			\toprule
			Attack Methods& R-1\tnote{1}~$\uparrow$  & R-2$\uparrow$   & R-3$\uparrow$   & R-5$\uparrow$   & R-10$\uparrow$  & FID$\downarrow$    & $d_{\rm MM}\downarrow$ & PPL\tnote{2}~$\downarrow$      & AS\tnote{3}$\downarrow$  \\
			\midrule
			\multicolumn{1}{l}{MLD}\\
			~~~~~~Target Motion    & 8 / 20  & 10 / 20          & 13 / 20          & 15 / 20          & 16 / 20          & 4.015          & 4.029 & 391.055  & -\\
			~~~~~~MacPromp         & 5 / 20 & 8 / 20           & 10 / 20 & 13 / 20 & 15 / 20          & 13.935         & 6.534          & 3061.488         & 0.471                   \\
			~~~~~~RIATIG           & 4 / 20          & 7 / 20           & \textbf{11 / 20}           & 13 / 20 & 17 / 20 & 10.899          & 5.368          & 1102.100
			& 0.131                   \\
			~~~~~~ALERT-Motion & \textbf{6 / 20} & \textbf{9 / 20}  & 9 / 20 & \textbf{15 / 20} & \textbf{19 / 20}          & \textbf{8.881} & \textbf{5.016} & \textbf{113.223}  & \textbf{0.067}          \\
			\midrule
			\multicolumn{1}{l}{MDM (100 steps)}\\
			~~~~~~Target Motion    & 7 / 20          & 14 / 20          & 15 / 20          & 16 / 20          & 20 / 20          & 4.055          & 3.549          & 391.055          & -                   \\
			~~~~~~MacPromp         & \textbf{7 / 20} & 7 / 20           & 8 / 20           & 16 / 20          & 16 / 20          & 11.108         & 5.106          & 2698.972         & 0.484                   \\
			~~~~~~RIATIG           & 5 / 20          & 8 / 20           & 10 / 20           & 16 / 20          & 18 / 20          & 12.435         & 5.024          & 1154.017          & 0.129                   \\
			~~~~~~ALERT-Motion & \textbf{7 / 20} & \textbf{13 / 20} & \textbf{14 / 20} & \textbf{17 / 20} & \textbf{19 / 20} & \textbf{5.843} & \textbf{4.117} & \textbf{179.496} & \textbf{0.075}          \\
			\midrule
			\multicolumn{1}{l}{MDM (1000 steps)} \\
			~~~~~~Target Motion    & 8 / 20          & 12 / 20          & 12 / 20          & 14 / 20          & 19 / 20          & 5.954          & 4.116          & 391.055          & -                   \\
			~~~~~~MacPromp         & 3 / 20          & 6 / 20           & 8 / 20           & 10 / 20          & 14 / 20          & 11.149         & 6.156          & 3023.887         & 0.467                   \\
			~~~~~~RIATIG           & 4 / 20          & 7 / 20           & 10 / 20           & \textbf{14 / 20}          & 16 / 20          & 9.875          & 5.444          & 1262.338          & 0.129                   \\
			~~~~~~ALERT-Motion & \textbf{9 / 20} & \textbf{12 / 20} & \textbf{13 / 20} & \textbf{14 / 20} & \textbf{19 / 20} & \textbf{6.183} & \textbf{4.533} & \textbf{140.793} & \textbf{0.074}          \\
			\bottomrule
		\end{tabular*}
\begin{tablenotes}  
\footnotesize 
\item[1] R-1, R-2, R-3, R-5, R-10 represent R-precision at R equals 1, 2, 3, 5, and 10, respectively. 
\item[2] PPL represents the perplexity of sentences.
\item[3] AS stands for Adversarial Similarity, which denotes the similarity between adversarial prompts and target prompts.
\end{tablenotes}
\end{threeparttable}
\end{table*}

\begin{table}[tp]
\centering
\caption{Attack performance on TMR evaluation model.}\label{tb:eval_model}
\resizebox{\columnwidth}{!}{%
	\begin{tabular}{llrrrrr}
		\toprule
		Attack Methods   & R-1$\uparrow$   & R-2$\uparrow$   & R-3$\uparrow$   & R-5$\uparrow$   & R-10$\uparrow$   \\
		\midrule
		\multirow{1}{*}{MLD} \\
		~~~~~~MacPromp         & 5 / 20          & 6 / 20           & 7 / 20           & 9 / 20           & 13 / 20           \\
		~~~~~~RIATIG           & 6 / 20          & 7 / 20           & 8 / 20           & 11 / 20           & \textbf{16 / 20}  \\
		~~~~~~ALERT-Motion & \textbf{8 / 20} & \textbf{9 / 20}  & \textbf{10 / 20} & \textbf{12 / 20} & 12 / 20  \\
		\midrule
		
		\multirow{1}{*}{MDM (100 steps)}\\
		~~~~~~MacPromp         & 4 / 20          & 6 / 20           & 9 / 20           & 11 / 20          & 16 / 20           \\
		~~~~~~RIATIG           & 5 / 20          & 6 / 20           & 7 / 20           & 11 / 20 & 14 / 20  \\
		~~~~~~ALERT-Motion & \textbf{7 / 20} & \textbf{12 / 20} & \textbf{15 / 20} & \textbf{16 / 20} & \textbf{17 / 20}           \\
		
		\midrule
		\multirow{1}{*}{MDM (1000 steps)}\\
		~~~~~~MacPromp         & 3 / 20          & 6 / 20           & 9 / 20           & 10 / 20          & 15 / 20           \\
		~~~~~~RIATIG           & 3 / 20          & 7 / 20           & 11 / 20           & 12 / 20           & 16 / 20           \\
		~~~~~~ALERT-Motion & \textbf{6 / 20} & \textbf{11 / 20}  & \textbf{12 / 20} & \textbf{14 / 20} & \textbf{18 / 20}  \\
		
		\bottomrule
	\end{tabular}}
\end{table}
\subsubsection*{Evaluation Setup.}
In our experiments, all attacks are conducted in a black-box setting, meaning that we generate motion only by querying the model with prompts and obtaining the generated results. We utilize the ``gpt-3.5-turbo-instruct'' API with ChatGPT to implement our approach. The initial adversarial prompt text is a motion description randomly generated by ChatGPT. We set the number of iterations as $50$, the size of the prompt set as $20$. We set the similarity threshold $\eta$ as $0.4$. Examples for the attack were taken from the top $20$ of the Dissimilar subset in the evaluation setup of~\cite{petrovich2023tmr}, where the model achieves the highest accuracy. During the attack process, we use the model from~\cite{petrovich2023tmr} to extract the motion features to compute cosine similarity and adopt the text feature extraction from~\cite{yang-etal-2021-universal}. The effectiveness is evaluated using T2M~\cite{guo2022generating}.
\begin{table*}[tp]
	\centering
	\caption{The mean and variance of evaluation metrics under different selections of target motion.}
	\label{tb:mean_var}
	\begin{tabular*}{\textwidth}{@{\extracolsep{\fill}\,}lrrrr}
		\toprule
		Metrics & Target Motion   & MacPromp        & RIATIG          & ALERT-Motion   \\
		\midrule
		R-1$\uparrow$     & (10.80 $\pm$1.94) / 20 & (4.00 $\pm$2.00) / 20  & (4.40 $\pm$1.02) / 20  & \textbf{(5.40 $\pm$1.36) / 20}   \\
		R-2$\uparrow$     & (13.20 $\pm$2.32) / 20 & (6.40 $\pm$3.38) / 20  & (6.40 $\pm$0.80) / 20  & \textbf{(8.20  $\pm$1.47) / 20}   \\
		R-3$\uparrow$     & (15.20 $\pm$2.32) / 20 & (8.80 $\pm$2.99) / 20  & (8.40 $\pm$1.50) / 20  & \textbf{(10.00 $\pm$0.89) / 20}   \\
		R-5$\uparrow$     & (17.00 $\pm$2.00) / 20 & (13.00 $\pm$1.90) / 20 & (13.20 $\pm$0.40) / 20 & \textbf{(13.40 $\pm$2.42) / 20}  \\
		R-10$\uparrow$    & (19.00 $\pm$0.63) / 20 & (17.20 $\pm$1.17) / 20 & (17.00 $\pm$1.62) / 20 & \textbf{(17.60 $\pm$1.02) / 20}  \\
		FID$\downarrow$     & 2.99$\pm$0.89       & 15.40$\pm$4.12      & 12.82$\pm$1.72      & \textbf{8.20$\pm$2.48}       \\
		$d_{\rm MM}\downarrow$ & 3.46$\pm$0.54       & 5.99$\pm$0.47       & 6.10$\pm$0.59       & \textbf{5.68$\pm$0.76}        \\
		PPL$\downarrow$     & 327.83$\pm$128.34   & 2571.15$\pm$397.28  & 1389.67$\pm$373.37   & \textbf{119.58$\pm$10.54}      \\
		AS$\downarrow$      & -~~~~~~~~~~       & 0.49$\pm$0.02       & 0.12$\pm$0.01       & \textbf{0.08$\pm$0.01}        \\
		\bottomrule
	\end{tabular*}
\end{table*}

\subsubsection*{Baselines.}
To the best of our knowledge, there is currently no targeted adversarial attack specifically designed for T2M generation. For comparison, we select two state-of-the-art targeted adversarial attack methods for text-to-image generation, MacPromp~\cite{milliere2022adversarial} and RIATIG~\cite{liu2023riatig}, as baseline methods. Since their tasks do not involve motion, we modify their task settings to match our task.

\subsection{Evaluation Metrics}
\subsubsection*{Motion Performance.}
We utilize the R Precision, a widely-used metric in T2M~\cite{guo2022tm2t,zhang2023t2m,tevet2022human,chen2023executing} to evaluate generated motion. This assessment involves comparing each motion not only with its ground-truth text but also with a misaligned description. R Precision is determined through the Euclidean distance between motion and text features. Our evaluation centers on measuring the average accuracy among the top-$k$ ranked descriptions. A ground-truth within the top-$k$ candidates is considered a ``True Positive'' retrieval. Our approach involves a batch size of 20, encompassing 19 negative examples, and we explore the effectiveness of R at various values: 1 (R-1), 2 (R-2), 3 (R-3), 5 (R-5), and 10 (R-10). Furthermore, our study incorporates Frechet Inception Distance (FID) as a metric to assess the quality of generated motion. FID, a widely accepted standard for evaluating content quality~\cite{tevet2022human,chen2023executing}, involves comparing features extracted from generated motion and real motion. In our motion domain adaptation, we adopt an evaluator network to represent deep features, deviating from the original image-based Inception neural network. Smaller FID values are indicative of superior results. Additionally, we compute the Multimodal Distance ($d_{\rm MM}$), which is the mean Euclidean distance between the motions features and their corresponding textual descriptions features in the test examples~\cite{tevet2022human,chen2023executing}. A lower value indicates better alignment between prompts and their generated motions.
\subsubsection*{Adversarial Similarity.}

\begin{table*}[tp]
	\centering
	\caption{The attack performance of 100 additional experiments.}
	\label{tb:100_add}
	\begin{tabular*}{\textwidth}{@{\extracolsep{\fill}\,}lccccccccc}
		\toprule
		Attack Methods & R-1$\uparrow$  & R-2$\uparrow$   & R-3$\uparrow$   & R-5$\uparrow$   & R-10$\uparrow$  & FID$\downarrow$    & $d_{\rm MM}$$\downarrow$ & PPL$\downarrow$      & AS$\downarrow$  \\
		\midrule
		Target Motion    & 54 / 100 & 66 / 100 & 76 / 100 & 85 / 100 & 95 / 100 & 0.92 & 3.46    & 327.83 & -       \\
		\midrule
		MacPromp & 20 / 100 & 32 / 100 & 44 / 100 & 65 / 100 & 86 / 100 & 8.37 & 5.99 & 2571.15 & 0.49  \\
		RIATIG   & 22 / 100 & 32 / 100 & 42 / 100 & 66 / 100 & 85 / 100 & 7.41 & 6.10 & 1389.67  & 0.12 \\
		ALERT-Motion & \textbf{27 / 100} & \textbf{41 / 100} & \textbf{50 / 100} & \textbf{67 / 100} & \textbf{88 / 100} & \textbf{4.17} & \textbf{5.68}    & \textbf{119.58}  & \textbf{0.08}     \\
		\bottomrule
	\end{tabular*}
\end{table*}
In adversarial attacks, it is essential for adversarial prompt  text to have low similarity with the target prompt text to evade detection. In line with previous studies, our initial step involves utilizing the Universal Sentence Encoder~\cite{cer2018universal} for encoding both the adversarial sentence and the target sentence, resulting in high-dimensional vectors. Subsequently, we determine their adversarial similarity by computing the cosine score.
\subsubsection*{Naturality.}
To ensure the naturalness of adversarial examples, we measure perplexity (PPL) using GPT-2~\cite{radford2019language}, trained on real-world sentences. PPL assesses the likelihood of the model in generating the input text, thereby indicating natural fluency of the adversarial prompts. Lower PPL values typically signify higher naturalness.
\subsection{Evaluation Results}
The attack results on MLD and MDM are shown in Table~\ref{tb:all}. Compared to the baselines, ALERT-Motion achieves a higher R-precision. Although MacPromp achieves higher R-precision and lower FID and $d_{\rm MM}$ in some cases, its direct translation of target prompts in various languages results in unnatural adversarial prompts. The perplexity is much higher than other methods, and, on the other hand, it closely resembles the target sentences, resulting in high adversarial similarity, making it less practical. RIATIG, compared to MacPromp, achieves similar or even higher R-precision, with a slight decrease in perplexity and adversarial similarity. However, as seen in Fig.~\ref{fig:MLD_examples}, there are still incorrect words and some extra spaces.

From Table~\ref{tb:all}, it can be observed that our proposed ALERT-Motion performs better on most metrics across these models. Additionally, examining Figs.~\ref{fig:MDM_examples} and \ref{fig:MLD_examples}, the adversarial prompts generated by ALERT-Motion are not only more natural but also relevant to the motion. In contrast, prompts obtained by other methods are mostly irrelevant to motion.
\subsection{Ablation Study}
\subsubsection*{Influence of Evaluation Models.}
The current research on the evaluation of motion generation is still limited, with T2M~\cite{guo2022generating} being widely recognized. Studies on motion generation, such as~\cite{zhang2023t2m,tevet2022human}, and~\cite{chen2023executing}, adopt T2M to assess the quality of generated models. The latest research on the evaluation of motion generation is presented in TMR ~\cite{petrovich2023tmr}. Therefore, we also use it to evaluate our experiments. As shown in Table~\ref{tb:eval_model}, under TMR model, ALERT-Motion exhibits significant superiority compared to other baseline methods, indicating that the excellent performance of our proposed ALERT-Motion is not influenced by the choice of evaluation models.
\subsubsection*{Influence of Target Motion. }
To further analyze the selected target motion on the attack performance, we chose all 100 motion from the Dissimilar subset evaluation setting in TMR~\cite{petrovich2023tmr} as target motion. We conduct five experiments, each using a different set of 20 target motion, following the order specified in their setting. Table~\ref{tb:mean_var} demonstrate that the superiority of ALERT-Motion performance over baseline methods remains consistent across different target motion. The overall performance of the attack method on these 100 target motion is presented in Table~\ref{tb:100_add}, demonstrating the consistent performance of our attack method.
\section{Discussion}
\subsubsection*{Safety Concerns.} To prevent potential malicious uses, it is crucial to consider defense mechanisms for T2M models. There are valid concerns around malicious actors exploiting such models to produce explicit violent content~\cite{birhane2021multimodal,yu2022scaling}. Moreover, considering that these models serve as humanoid controllers~\cite{luo2023perpetual} and may potentially be utilized for humanoid robots in the future, there are risks of the robots behaving in ways that endanger humans if not properly constrained. Although there is currently no research specifically addressing defense mechanisms for T2M generation models, we demonstrates that existing content moderation filters, if directly deployed on T2M models, are bypassed by our adversarial attack method. Moreover, the fact that our method creates adversarial prompts related to T2M task makes the attacks more challenging to defend against. Therefore, the safety risks of using motion generation models must be taken into consideration.
\subsubsection*{Potential Defense Strategies.} Several defense methods may be considered. Rule-based text filters might find our approach challenging to counter since the adversarial prompts we create seamlessly blend into normal text, remaining semantically related to motion. One potential defense mechanism involves leveraging larger datasets for training. The most extensive dataset in current motion generation research~\cite{guo2022generating} comprises just over 10,000 text-motion pairs. We hypothesize that increasing the volume of training data boosts the alignment between the generated model and T2M tasks. Furthermore, established defense methods from the realms of adversarial attacks and NLP~\cite{goodfellow2014explaining,jin2020bert} is applied to strengthen T2M models, specifically through adversarial training.
\section{Conclusion}
In this paper, a novel method that involves conducting targeted adversarial attack against T2M models is proposed. Additionally, we introduce an autonomous LLM-enhanced adversarial attack method called ALERT-Motion, which comprises two modules: the multimodal information integration (MMIC) module and the adaptive dispatching (AD) module. Assisted by MMIC, AD, with the incorporation of LLM during progresses of expansion, refinement, and updating, autonomously learns and executes the search for optimal adversarial prompts. Our extensive experiments validate the ability to discover adversarial prompts that exhibit both fluency and related to motion. Moreover, these prompts trigger the victim T2M model to generate motion closely resembling the target, thus achieving successful attacks. The susceptibility of T2M models to our attacks suggest an urgent need to develop defensive methods and improve the robustness against adversarial exploitation.

{
    \small
    \bibliographystyle{ieeenat_fullname}
    \bibliography{main}
}


\end{document}